\pdfoutput=1

\documentclass[11pt,a4paper]{article}
\usepackage[hyperref,final]{acl}
\usepackage{times}
\usepackage{inconsolata}
\usepackage{latexsym}
\usepackage{danudefs}
\usepackage{algorithmic}
\usepackage{algorithm}
\usepackage{color}
\usepackage{booktabs}
\usepackage{xspace}
\usepackage{url}
\usepackage{braket}
\usepackage[T1]{fontenc}
\usepackage[utf8]{inputenc}
\usepackage{graphicx}
\usepackage{microtype}
\usepackage[table]{xcolor}
\usepackage{graphicx}
\usepackage{multirow}
\usepackage{colortbl}
\usepackage{array}
\usepackage{booktabs}
\usepackage{tcolorbox}
\usepackage{enumitem}
\tcbuselibrary{breakable, skins}
\usepackage{pifont}

\usepackage[english]{babel}
\usepackage{hyperref}
\addto\captionsenglish{%
}
\addto\extrasenglish{%
}

\usepackage{acronym}

\usepackage{etoolbox}
\makeatletter
\newif\if@in@acrolist
\AtBeginEnvironment{acronym}{\@in@acrolisttrue}
\newrobustcmd{\LU}[2]{\if@in@acrolist#1\else#2\fi}

\newcommand{\ACF}[1]{{\@in@acrolisttrue\acf{#1}}}
\makeatother

\acrodef{NLP}[NLP]{Natural Language Processing}
\acrodef{MLM}[MLM]{Masked Language Model}
\acrodefplural{MLM}[MLMs]{\LU{M}{m}asked \LU{L}{l}anguage \LU{M}{m}odels}
\acrodef{LLM}[LLM]{Large Language Model}
\acrodefplural{LLM}[LLMs]{Large Language Models}
\acrodef{SoTA}[SoTA]{state-of-the-art}
\acrodef{ICL}{\LU{I}{i}n-cotext \LU{L}{l}earning}
\acrodef{SCD}{\LU{S}{s}emantic \LU{C}{c}hange \LU{D}{d}etection}
\acrodef{WiC}{Word-in-Context}
\acrodef{ITML}[ITML]{Information-Theoretic Metric Learning}
\acrodef{SDML}[SDML]{Semantic Distance Metric Learning}
\acrodef{MLP}[MLP]{Multi-layer Perception}
\acrodef{FFN}[FFN]{Feed Forward Network}

\acrodef{SCD}{\LU{S}{s}emantic \LU{C}{c}hange \LU{D}{d}etection}
\acrodef{WiC}{Word-in-Context}
\acrodef{C-STS}[C-STS]{Conditional Semantic Textual Similarity}
\acrodef{STS}[STS]{Semantic Textual Similarity}
\acrodef{KGC}[KGC]{Knowledge Graph Completion}

\acrodef{MTEB}[MTEB]{Massive Text Embedding Benchmark}

\acrodef{NV}[NV]{NV-Embed-v2}
\acrodef{SFR}[SFR]{SFR-Embedding-Mistral}
\acrodef{GTE}[GTE]{gte-Qwen2-7B-instruct}
\acrodef{E5}[E5]{Multilingual-E5-large-instruct}
\acrodef{SimCSE-large}[SimCSE\_large]{sup-simcse-roberta-large}
\acrodef{SimCSE-base}[SimCSE\_base]{sup-simcse-bert-base-uncased}

\title{Annotating Training Data for Conditional Semantic Textual Similarity Measurement using Large Language Models}

\author{
Gaifan Zhang\textsuperscript{1} \quad
Yi Zhou\textsuperscript{2} \quad
Danushka Bollegala\textsuperscript{1} \\
\textsuperscript{1} University of Liverpool \quad
\textsuperscript{2} Cardiff University \\
\texttt{sggzhan8@liverpool.ac.uk, zhouy131@cardiff.ac.uk, danushka@liverpool.ac.uk}
}

\date{}

\begin{document}
\maketitle

\begin{abstract}
    Semantic similarity between two sentences depends on the aspects considered between those sentences.
    To study this phenomenon, \citet{Deshpande:2023} proposed the \ac{C-STS} task and annotated a human-rated similarity dataset containing pairs of sentences compared under two different conditions.
    However, \citet{L-CSTS} found various annotation issues in this dataset and showed that manually re-annotating a small portion of it leads to more accurate \ac{C-STS} models.
    Despite these pioneering efforts, the lack of large and accurately annotated \ac{C-STS} datasets remains a blocker for making progress on this task as evidenced by the subpar performance of the \ac{C-STS} models.
    To address this training data need, we resort to \acp{LLM} to correct the condition statements and similarity ratings in the original dataset proposed by \citet{Deshpande:2023}.
    Our proposed method is able to re-annotate a large training dataset for the \ac{C-STS} task with minimal manual effort.
    Importantly,  by training a supervised \ac{C-STS} model on our cleaned and re-annotated dataset, we achieve a 5.4\% statistically significant improvement in Spearman correlation.
    The re-annotated dataset is available at \url{https://LivNLP.github.io/CSTS-reannotation}.
\end{abstract}

\section{Introduction}
\label{sec:intro}

\begin{figure}
    \centering
    \includegraphics[width=1.0\linewidth]{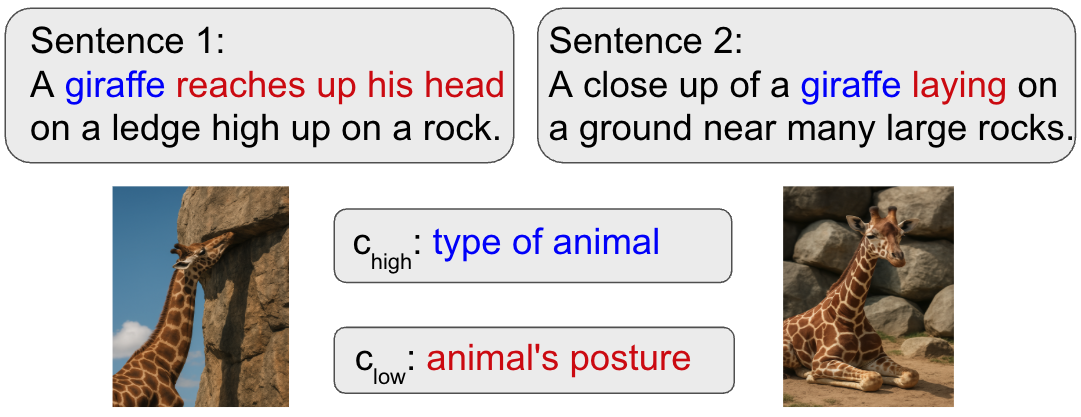}
    \caption{An example \ac{C-STS} instance. The two sentences are compared under two different conditions, focusing on different aspects, resulting in a high (score of 5), and a lower (score of 1) semantic similarities. Images are only for visual cue.}
    \label{fig:intro}
\end{figure}

\ac{STS} is a fundamental \ac{NLP} task to evaluate the semantic similarity between two given sentences \cite{Agirre:2012}. 
However, the focus on the sentences can vary and affects the judgment of similarity. 
To address this, \citet{Deshpande:2023} introduced a novel \ac{C-STS} task, which measures the similarity between two sentences under a specified condition.
In the \ac{C-STS} dataset, each sentence pair has two conditions -- a condition $c_{\rm low}$ producing a low semantic similarity, and a condition $c_{\rm high}$ a high semantic similarity, as shown in \autoref{fig:intro}.
The similarity under each condition is rated on an ordinal scale from 1 (low similarity) to 5 (high similarity).

While the C-STS task brings greater specificity to the aspects of sentences being compared, \citet{L-CSTS} observed that both the conditions and human similarity ratings suffer from issues such as ambiguity and inaccuracy, introducing label noise into the task.
Although recent methods \cite{li2024seaver, liu2025conditional, Hyper-CL} have advanced the modeling of \ac{C-STS}, their performance is still limited by the dataset quality, with Spearman correlations generally remaining below 0.5.
To reduce those identified annotation errors, \citet{L-CSTS} re-annotated the validation portion of the dataset with the help of human annotators.
However, as discussed later in \autoref{sec:cond-modi}, in addition to annotation errors in similarity ratings, we find that the conditions themselves can be problematic, such as expressing varying granularities and a high-level of subjectivity, further impacting the reliability of the dataset.
Moreover, the validation data re-annotated by \citet{L-CSTS} consists of only a small proportion (15\%) of the \ac{C-STS} dataset.
Although it would be ideal to manually re-annotate the full \ac{C-STS} dataset it is a costly task.

To address this data cleansing task, we use \acp{LLM} to 
(1) modify the conditions, and 
(2) re-annotate the similarity ratings between two sentences under the modified conditions, requiring minimum manual effort.
\acp{LLM} have been successfully used to generate synthetic training data and to provide judgements for several related NLP tasks~\citep{peng2023generating, patel2024datadreamer, wei2024systematic}.
It is noteworthy that prior work \cite{Deshpande:2023} using \acp{LLM} such as GPT-4~\citep{OpenAI:2023} and Flan-T5~\citep{T5} to predict \ac{C-STS} have reported suboptimal performance where they observed numerous issues including semantically similar sentence pairs being incorrectly assigned with low similarity scores.
While we also use \acp{LLM} to correct the conditions and similarity ratings, we aim to improve the effectiveness of the \ac{C-STS} training data by improving annotation accuracy and increasing the number of high-quality and reliable instances, such that better \ac{C-STS} models can be trained.

\begin{table}[t]
\centering
\resizebox{\linewidth}{!}{%
\begin{tabular}{l r r r r }
\textbf{Dataset} & \textbf{Train} & \textbf{Validation} & \textbf{Test} & \textbf{Count}\\
\toprule
\citet{Deshpande:2023} & 11342 & 2834 & 4732 & 18908\\
\citet{L-CSTS} &  &\ding{51} & & 2834\\
Ours & \ding{51} &\ding{51} & & \textbf{14176}\\
\bottomrule
\end{tabular}
}
\caption{Dataset size comparison. The portions that have been re-annotated by Tu et al. and this work (ours) are indicated by \ding{51}.}
\label{tab:dataset_size}
\end{table}

We cleaned the training dataset proposed by \citet{Deshpande:2023}, which accounts for 75\% of the whole dataset, as demonstrated in \autoref{tab:dataset_size}.
This provides more reliable training instances for the \ac{C-STS} task.
Since the test set labels have not been released, we do not modify the test instances.
Our contributions in this paper are three-fold.
\begin{enumerate}
    \item We first correct the errors and refine the expressions in the condition statements in the \ac{C-STS} dataset (\autoref{sec:cond-modi}).
    \item Next, we use two \acp{LLM} (i.e. \texttt{GPT-4o} and \texttt{Claude-3.7-Sonnet}) to independently obtain \ac{C-STS} ratings, which we then combine with the original human ratings by averaging (\autoref{sec:reanno}). 
    \item To evaluate the usefulness of our \ac{LLM}-cleansed dataset, we train a supervised \ac{C-STS} model on it following the method proposed by \citet{zhang2025caseconditionawaresentence}.
\end{enumerate}
Our evaluations show that the trained model obtains a Spearman correlation of 73.9\% against the human-rated test data, thereby demonstrating the usefulness of our dataset when training \ac{C-STS} models.
Specifically, our cleaned and re-annotated dataset achieves a 5.4\% statistically significant improvement measured in Spearman correlation.

\section{C-STS Training Data Cleansing}
\label{sec:cur}

Our data cleansing method for \ac{C-STS} consists of two steps. 
In the first step (\autoref{sec:cond-modi}), we identify common issues with the conditions and use \texttt{GPT-4o} to refine those. 
In the second step (\autoref{sec:reanno}), we re-annotate the labels using both \texttt{GPT-4o} and \texttt{Claude-3.7-Sonnet}, due to their high performance on natural language understanding as demonstrated by Chatbot Arena leader-board \cite{zheng2023judgingllmasajudgemtbenchchatbot}.\footnote{\url{https://lmarena.ai/}}
Empirically, we find that both of those \acp{LLM} generated ratings demonstrate a high level of agreement with the human \ac{C-STS} ratings, resulting in Spearman correlations of 62\% and 66\% on the human-reannotated test set (ReTest) by \citet{L-CSTS}, respectively.
Finally, we aggregate the human ratings in the original dataset with the two sets of \ac{LLM} ratings.

\subsection{Modifying the Conditions}
\label{sec:cond-modi}

\begin{table}[!t]
\centering
\renewcommand{\arraystretch}{1.3}
\resizebox{0.9\linewidth}{!}{
\begin{tabular}{p{0.13\textwidth}|p{0.32\textwidth}}
\textbf{Issue} & \textbf{Condition} \\
\toprule
Imbalanced Condition & number of \# \newline
type of \# \newline
color of \# \\
\hline
Subjective Condition &  The age of person. \newline
The color of animal. \newline
The number of people. \\
\hline
Inconsistent Phrasing Style & 
The all are food. \newline
Where the dog is visible from. \newline
The amount of stoves/ ovens. \newline
Type of room. \newline
The person's age. \\
\hline
Varying Granularity & The absence of tomato. \newline
The place of the object. \newline
The species of the one who's in the room. \\
\hline
Verbose Expression & The fact that they're both girls. \newline
String instrument being played. \newline
The players move to the position. \\
\hline
Grammatical Issue & The thing that fly. \\
\bottomrule
\end{tabular}
}
\caption{Common stand-alone condition issues.}
\label{tab:condition_only_issues}
\end{table}

\begin{table*}
\centering
\renewcommand{\arraystretch}{1.3} 
\resizebox{0.9\textwidth}{!}{
\begin{tabular}{p{0.21\textwidth}|p{0.6\textwidth}|p{0.14\textwidth}}
\textbf{Issue} & \textbf{Sentence Pair} & \textbf{Condition} \\
\toprule
Ambiguous Condition & A climber with a yellow backpack walks along the ridge of a snowy mountainside. \newline
A person in a red hat with a huge backpack going hiking. & The climber. \\
\hline
Invalid Condition & A man wearing yellow and blue is riding a large, bucking bull. \newline
A bull rider, in full padding and wearing a helmet, rides a large brown and white bull. & Color of bull. \\
\hline
Unrelated Condition & Three \underline{hotdogs} on buns with whole slices of relish sit on a white plate. \newline
A \underline{hot dog} on a bun with a drop of ketchup on the table. & The number of \underline{dogs}. \\[1ex]
\bottomrule
\end{tabular}
}
\caption{Common condition issues that cause the judgment divergence related to sentences.}
\label{tab:condition_sentence_issues}
\end{table*}

We identify multiple issues in the conditions that impact the accuracy of the human annotations. 
These issues fall into two categories: 
(1) conditions that are inherently ambiguous or misleading in their own (\textbf{stand-alone} condition issues),
and (2) conditions that are misleading when interpretting the sentence semantics (\textbf{sentence-dependent} condition issues).
Next, we describe those issues.

\subsubsection{Stand-alone Condition Issues}
\label{sec:standalone-cond}

\paragraph{Imbalanced Conditions:} Certain condition types occur far more frequently than the others, resulting in a highly imbalanced distribution (see \autoref{imbalanced_condition}), biasing model training and evaluation.
For example, the condition types \textit{number of \#} and \textit{type of \#} take 16.7\% and 16.6\% of the dataset, respectively.

\paragraph{Subjective Conditions:} Some conditions introduce discrepancies with the human similarity ratings because different annotators can interpret the same condition differently. 
As a result, different annotators can assign contradicting similarity ratings to the same sentence pair.
For example, when comparing the two numbers 2 and 3 (in the case of condition \textit{number of \#}), one annotator might consider the numerical closeness (i.e. 2 is closer to 3) as an indication of high similarity, while another may regard this as an inequality (i.e. 2 is not equal to 3), assigning a low similarity.
\autoref{subjective-label} presents examples of such subjectivity and inconsistency in human similarity judgments.
This annotation noise in the original \ac{C-STS} dataset reduces the reliability of model evaluations.

\paragraph{Inconsistent Phrasing Styles:} The phrasing of some conditions is inconsistent, ranging from full sentences to fragmented sentences or phrases.
Moreover, they lack uniformity in both stopword usage and their grammatical structure.

\paragraph{Varying Granularity:} Conditions range from very general to overly specific. 
This divergence affects how the models interpret those conditions.

\paragraph{Verbose Expressions:} Conditions can sometimes have over-complex expressions, including words that overly elaborate sentence structures. 

\paragraph{Grammatical Issues:} Obvious English grammatical errors exist in some of the conditions.

\autoref{tab:condition_only_issues} shows examples of the above-mentioned issues.

\subsubsection{Sentence-dependent Condition Issues}
\label{sec:sentence-cond}

\paragraph{Ambiguous Conditions:} \citet{L-CSTS} found that conditions presented as singletons without associated entity features to be ambiguous, lacking a clear specification of the aspects being compared.

\paragraph{Invalid Conditions:} \citet{L-CSTS} showed that some of the conditions to be invalid, as they require information that cannot be inferred from the sentences based on those conditions.

\paragraph{Unrelated Conditions:} Some conditions contain typos or imprecise expressions. Although comprehensible by humans, such issues could mislead embedding model judges.

\autoref{tab:condition_sentence_issues} shows examples of the above-mentioned issues.
We also observe overlaps of sentences and conditions between the training and test sets (see \autoref{overlapping} for details), which can overestimate the generalisability of the models.
To standardise the condition expressions and improve their specificity and accuracy to reduce ambiguity, we use \texttt{GPT-4o} to refine the conditions.
The complete prompt, along with examples before/after the modified conditions, is provided in \autoref{cond-modification}.
Specifically, we instruct \texttt{GPT-4o} using a prompt that provides explicit guidelines and constraints. 
The prompt requires that conditions to be clear, specific, and semantically grounded, discouraging vague references (e.g., ``animal'') in favour of more precise formulations (e.g., ``species of animal''). 
We also remove redundant stopwords (e.g., ``the'') and maintain a uniform phrasing style across all conditions. 
Additionally, the prompt requests a justification for any substantive modifications.

\subsection{Re-annotating the Similarity Ratings}
\label{sec:reanno}

After refining the conditions, we use \acp{LLM} to re-annotate the similarity ratings in the training set.
Specifically, we use \texttt{GPT-4o} and \texttt{Claude-3.7-Sonnet} with a few-shot prompt, providing five examples covering similarity ratings (1–5), each accompanied by a human-written justification. 
We also require \acp{LLM} to give corresponding justifications for their similarity ratings.
This design serves two purposes: 
(1) it helps the \ac{LLM} to understand the scoring rubric in a conditional \ac{STS} context;
and (2) it encourages the generation of not only a similarity rating but also a justification, which serves as a self-check mechanism to reduce hallucinations and improve the annotation quality. 
We use the same five-point rating scale proposed by \citet{Deshpande:2023} and instruct the \acp{LLM} to only return a JSON-formatted object instead of a natural language commentary.
The complete prompt, along with examples before/after re-annotating the similarity ratings under the modified conditions is provided in \autoref{label-reanno}.

Our preliminary analysis of the condition patterns and human ratings showed that the condition type \textit{number of \#} takes the largest proportion in the dataset and has a serious problem of subjectivity as described in \autoref{sec:standalone-cond}.
Therefore, we provide additional clarification and instructions to \acp{LLM} along with the general scoring definition by \citet{Deshpande:2023}.
We adopt the re-annotation strategy of \citet{L-CSTS}, assigning high similarity scores to sentence pairs that contain the same counted number and low similarity scores when the numbers differ.
If the numbers cannot be counted explicitly, the annotation relies on the approximate quantities and follows the general similarity definition.
This adjustment improves the consistency and interpretability of the dataset on this specific condition type.

To further increase the reliability of the annotations, we combine the original human ratings with multiple \ac{LLM}-predicted ratings.
Specifically, for each instance, we compute the arithmetic mean of the original human-annotated similarity rating ($y^{\text{human}}$), the predicted ratings by \texttt{GPT-4o} ($y^{\text{GPT-4o}}$), and  \texttt{Claude-3.7-Sonnet} ($y^{\text{Claude}}$), and round the result to the nearest integer. 
As shown in \autoref{voting}, combining ratings from both \acp{LLM} results in the best performance.

\section{Experiments}
\label{sec:exp}

For ease of disposition, we define the following dataset naming conventions.
\textbf{Train-Orig} is the original training set from \citet{Deshpande:2023}. \textbf{Train-Mod} applies condition modifications to Train-Orig, and \textbf{Train-Mod-Reanno} further includes our re-annotated ratings.
\textbf{Val-Orig} denotes the original validation set, and \textbf{Val-Reanno} is the \emph{human} re-annotated version introduced by \citet{L-CSTS}. 
Val-Reanno is the most accurate human-verified \ac{C-STS} data to date.
We split \textbf{Val-Reanno} into \textbf{ReVal} (randomly selected 70\%) as our validation set and \textbf{ReTest} (remaining 30\%) as our test set.
We construct \textbf{ReVal-Mod} and \textbf{ReTest-Mod} by applying condition modifications to ReVal and ReTest, respectively.

To evaluate the effectiveness of a particular training dataset, we first use it to train a supervised Non-Linear Projection (\textbf{SNPro}) model following \citet{zhang2025caseconditionawaresentence}, and then measure the improvement of \ac{C-STS} task performance on the same human-labelled test data (ReTest).
Details of this supervised model architecture are provided in \autoref{multi-head}.
Spearman's correlation coefficient with human similarity ratings is the standard evaluation metric for \ac{C-STS}, where a high correlation indicates an accurate \ac{C-STS} model.
We use an NVIDIA RTX A6000 GPU with PyTorch 2.0.1 and CUDA 11.7 for our experiments.

\begin{table}[t]
\centering
\resizebox{0.9\linewidth}{!}{%
\begin{tabular}{llc}
\toprule
\textbf{Train} & \textbf{Test}   & \textbf{Spearman} \\
\hline
ReVal & ReTest & 61.28 \\
ReVal-Mod w/o & ReTest-Mod w/o   & 64.25 \\
ReVal-Mod w/ & ReTest-Mod w/ & \textbf{66.89} \\
\bottomrule
\end{tabular}
}
\caption{Comparison of condition modification, evaluated using an SNPro model.
\textit{w/} and \textit{w/o}  denote condition modification with and without stopword removal, respectively.}
\label{tab:cond-mofi-validation}
\end{table}

To evaluate the effectiveness of condition modification, we train \textbf{SNPro} models on ReVal and evaluate on ReTest as shown in \autoref{tab:cond-mofi-validation}.
Further effect of stopword removal from the modified conditions is also considered.
We see that the best performance is reported by the \ac{LLM}-based condition modification with stopword removal (i.e. ReText-Mod w/).
Stopwords often contribute little or no semantic distinctions to the conditions, and removing them helps the model to attend to content words.

Following these findings, we apply condition modification with stopword removal and follow \autoref{sec:reanno} to re-annotate the similarity ratings in the condition-modified \ac{C-STS} training set. 
To measure the consistency of \ac{LLM}-generated annotations, 
we randomly select 100 instances from Train-Mod and repeat the annotation process five times using \texttt{Claude-3.7-Sonnet} with our few-shot prompt. 
We measured the agreement of the five sets of annotations using the Krippendorff’s Alpha~\cite{hayes2007answering} to be 0.865, indicating a high level of annotation consistency.

To validate the \ac{LLM}-modified conditions and re-annotated similarity ratings, we randomly selected 300 instances from our dataset to conduct a manual verification.
We find that the condition statements are clearer and more specific and in most condition statements, only stopwords are removed.
Importantly, we do not find any conditions that degrade in quality or meaning altered significantly.
On the other hand, we found that that 23\% (69/300) of the original human ratings to be inaccurate.
Roughly one-third of these inaccuracies involved serious errors, such as assigning high similarity scores to clearly dissimilar sentence pairs.

In contrast, investigating the re-annotated similarity ratings, we found that the re-annotated similarity ratings to accurately reflect the true conditional semantic textual similarity in most cases.
Cases where similar sentence pairs were previously labelled as dissimilar were correctly assigned higher similarity ratings during this re-annotation process.
A small proportion of instances (9\%, 28/300) deviate slightly from the human ratings, with a difference of only 1 point on the [1, 5] similarity scale.
Such minor disagreements are to be expected given the subjectivity involved in both the conditions and the meanings of the sentences.

\begin{figure}
    \centering
    \includegraphics[width=0.7\linewidth]{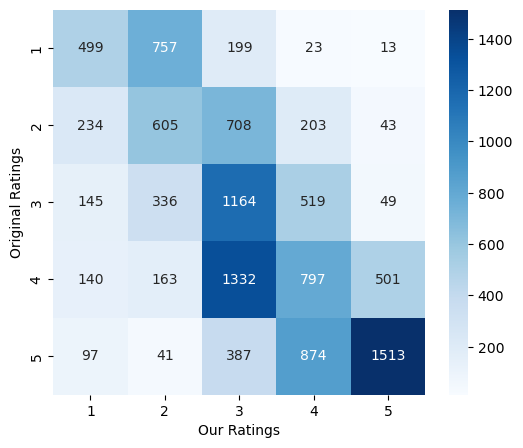}
    \caption{Confusion matrix between the original ratings (Train-Orig) and our re-annotated (Train-Mod-Reanno) ratings.}
    \label{fig:confusion_matrix}
\end{figure}

\autoref{fig:confusion_matrix} shows how our re-annotated ratings  (Train-Mod-Reanno) differ from the original annotations (Train-Orig).
Although there is a better agreement for high similarity annotations, we see less agreement for lower similarity ratings.
The relatively low Cohen's Kappa \cite{cohen1960coefficient} of 0.247 between the two sets of annotations indicates only fair agreement, highlighting that we have made significant revisions to the original \ac{C-STS} dataset.
Importantly, during our first step of modifying the conditions, we deliberately shifted the semantic focus of some sentences to ensure clearer, more consistent criteria.

\begin{table}[t]
\centering
\resizebox{0.9\linewidth}{!}{%
\begin{tabular}{llc}
\toprule
\textbf{Train} & \textbf{Test} & \textbf{Spearman} \\
\midrule
Train-Orig & ReTest &  68.54 \\
Train-Orig & ReTest-Mod & 69.68 \\
Train-Mod & ReTest-Mod & 69.39 \\
Train-Mod-Reanno & ReTest-Mod & \textbf{73.93} \\
\bottomrule
\end{tabular}
}
\caption{Spearman correlation coefficients obtained by training an SNPro model on different training datasets}
\label{tab:csts-performance}
\vspace{-4mm}
\end{table}

To evaluate the ability of our \ac{LLM} modified conditions and the re-annotated similarity ratings for improving \ac{C-STS} measurement, we train \textbf{SNPro} models using different training datasets in \autoref{tab:csts-performance}.
Compared to training \ac{C-STS} models on Train-Orig, we see that doing so on Train-Mod-Reanno results in the best performance.
This is a 5.4\% statistically significant improvement over the best bi-encoder \ac{C-STS} performance reported by \citet{zhang2025caseconditionawaresentence}.
This shows that, keeping the model architecture and all other training settings fixed, our re-annotated \ac{C-STS} training data alone can improve the performance of \ac{C-STS}.
We hope that our re-annotated \ac{C-STS} training data will expedite the future progress of \ac{C-STS} research.

\section{Conclusion}

We identify key issues in the condition definitions and human-annotated similarity ratings in the original \ac{C-STS} dataset. 
To address these, we propose an efficient \ac{LLM}-based data cleansing approach that improves dataset quality through condition modification and re-annotation of similarity scores. 
By integrating this with human-annotated data, our cleansed dataset significantly advanced the performance of a previously proposed \ac{C-STS} method.

\section{Limitations}

There is a large number of LLMs developed and made publicly available.
However, it is practically infeasible to use multiple LLMs for the C-STS data re-annotation due to the costs involved.
Therefore, we selected two highly popular and accurate models at the time of writing (\texttt{GPT-4o} and \texttt{Claude-3.7-Sonnet}) to balance performance and cost-effectiveness. 
Although we modified the conditions, certain stand-alone condition issues such as imbalanced conditions still exist, as the overall distribution of condition types has not changed.

This study was conducted using \ac{C-STS} datasets for English, which is a morphologically limited language.
However, this choice is based on the availability of \ac{C-STS} datasets.
To the best of our knowledge, \ac{C-STS} datasets are not publicly available for languages other than English.
We consider it to be an important task for future work to develop multilingual \ac{C-STS} datasets to study the language-specific issues pertaining to this task.

\section{Ethical Concerns}
\acp{LLM} have been shown to exhibit social biases, such as those related to age and gender \cite{gallegos-etal-2024-bias}.
Such social topics exist in the conditions for the \ac{C-STS} task.
Using \acp{LLM} for annotation may further propagate such biases into the dataset.
The influence of whether the \ac{LLM}-based annotation process impacts the data quality with respect to social bias is not evaluated in this study.
Additionally, \ac{LLM}-based condition-aware sentence embeddings could encode unfair social biases.
Therefore, it is important to evaluate social bias amplifications (if any) due to training \ac{C-STS} models on our proposed training dataset before deploying those models in downstream NLP applications.

\bibliography{csts}

\appendix
\section*{Supplementary Materials}
\section{Imbalanced Condition}
\label{imbalanced_condition}

By analysing the distribution of condition types in the \ac{C-STS} training dataset, we observe a significant imbalance.
As shown in \autoref{tab:general_condition_summary}, two broad condition categories, \textit{number of \#} and \textit{type of \#}, dominate the dataset, accounting for $16.7\%$ and $16.6\%$ of all conditions, respectively.

With respect to specific conditions, we present the 15 most frequent ones in \autoref{tab:specific_condition_count}.
The most common conditions include \textit{The number of people.}, \textit{The type of animal.}, and \textit{The sport.}
However, these frequently occurring conditions often introduce problems such as ambiguity and subjectivity in the evaluation process.

\begin{table}[ht]
\centering
\begin{tabular}{l r r}
\textbf{Condition Type} & \textbf{Count} & \textbf{Percentage} \\
\toprule
number of \# & 1892 & 16.7\% \\
type of \# & 1886 & 16.6\% \\
color of \# & 664 & 5.9\% \\
action & 357 & 3.1\% \\
position of \# & 88 & 0.8\% \\
\bottomrule
\end{tabular}
\caption{Counts of general condition types (top 5) in the original \ac{C-STS} training dataset.}
\label{tab:general_condition_summary}
\end{table}

\begin{table}[ht]
\centering
\begin{tabular}{l r}
\textbf{Condition} & \textbf{Count} \\
\toprule
The number of people. & 520 \\
The type of animal. & 254 \\
The sport. & 249 \\
The name of the place. & 162 \\
The animal. & 154 \\
The color of the shirts. & 123 \\
The number of people visible. & 103 \\
The action. & 94 \\
The type of food. & 87 \\
The number of animals. & 85 \\
The type of clothing. & 85 \\
The number of people in the image. & 72 \\
The location. & 65 \\
The color of the clothing. & 64 \\
The number of objects. & 62 \\
\bottomrule
\end{tabular}
\caption{Counts of specific conditions (top 15) in the original \ac{C-STS} training dataset.}
\label{tab:specific_condition_count}
\end{table}

\section{Subjectivity in Human Annotations}
\label{subjective-label}

Human annotators can give contradictory ratings to some similar instances in the dataset.
We show subjectivity in human ratings for the conditions \textit{The number of people}, \textit{Age of person} and \textit{Gender of person} in the original \ac{C-STS} training dataset as examples.
\autoref{tab:contrastive_subjective} lists some examples of instances that show subjectivity.
We explain them one by one as follows.

\begin{table*}[ht!]
\centering
\begin{tabular}{p{5cm} p{5cm} p{3.5cm} c}
\textbf{Sentence 1} & \textbf{Sentence 2} & \textbf{Condition} & \textbf{Rating} \\
\toprule
A man and woman sitting in a booth together and smiling. & Three people sitting at a table at a restaurant. & The number of people. & 4 \\
\hline
A baseball player swings to hit the ball as another player catches. & A man in a white and black uniform is attempting to swing a baseball bat. & The number of people. & 4 \\
\hline
A person is diving into blue water on a rocky coast. & Two males on a rock over water, one in midair about to dive. & The number of people. & 1 \\
\hline
A person is doing a trick in the air on a bike near some buildings. & Person performing a move on a mountain bike with two people watching. & The number of people. & 1 \\
\hline
A young girl with a sippy cup swings on a swing. & A child is making a ridiculous face with an open mouth. & The number of people. & 4 \\
\hline
The boy on the bike is wearing safety glasses and a red helmet. & A man dressed in bicycle gear is riding through a course. & Age of person. & 1 \\
\hline
Two images show a man reaching out to hit a tennis ball with a racket. & A boy in black shorts jumps and holds his tennis racket out in front of him. & Age of person. & 3 \\
\hline
A very happy child sits on a chair on top of some rocks. & A child is bouncing on a trampoline that is near a house. & Age of person. & 3 \\
\hline
A man in a red and yellow outfit is riding a bicycle on one wheel. & A woman is riding a bike with a basket of flowers. & Gender of person. & 1 \\
\hline
A woman with a red scarf around her neck is smiling. & A man in a black hat looks very happy. & Gender of person. & 4 \\
\hline
A little girl is brushing her teeth in a bathroom. & A woman is brushing her teeth in a bathroom mirror. & Gender of person. & 1 \\
\hline
A man is skateboarding on the sidewalk. & A girl is rollerblading on a path. & Gender of person. & 4 \\
\bottomrule
\end{tabular}
\caption{Examples of sentence pairs under the conditions ``The number of people'', ``Age of person'', and ``Gender of person'' with subjective similarity ratings by human annotators in the original \ac{C-STS} training set.}
\label{tab:contrastive_subjective}
\end{table*}

Considering the condition \textbf{\textit{The number of people}}:

In the instance that Sentence 1: \textit{A man and woman sitting in a
booth together and smiling.}, Sentence 2: \textit{Three people sitting at a table at
a restaurant.}, Rating: 4, there are 2 people in Sentence 1, and 3 people in Sentence 2. 
While the number of people differs (2 vs. 3), annotators still rated the pair as highly similar. 
This suggests that some annotators perceive small differences in number (such as 2 versus 3) as relatively minor.

In the instance that Sentence 1: \textit{A baseball player swings to hit
the ball as another player catches.}, Sentence 2: \textit{A man in a white and black uni-
form is attempting to swing a
baseball bat.}, Rating: 4, there are 2 people in Sentence 1 and 1 person in Sentence 2. 
Human annotators give this small difference in number a score of high similarity.

However, in another instance that Sentence 1: \textit{A person is diving into blue water
on a rocky coast.}, Sentence 2: \textit{Two males on a rock over water,
one in midair about to dive.}, Rating: 1, there are 1 person in Sentence 1 and 2 people in Sentence 2.
The number of people is also different, but similar in number (same case as the previous example). 
Some annotators interpret it as a stronger signal of dissimilarity.

Additionally, in the instance that Sentence 1: \textit{A person is doing a trick in the
air on a bike near some buildings.}, Sentence 2: \textit{Person performing a move on a
mountain bike with two people
watching.}, Rating: 1, there are 1 person in Sentence 1 and 3 people in Sentence 2. 
Human annotators can regard this mismatch in number as dissimilarity.

Moreover, in the instance that Sentence 1: \textit{A young girl with a sippy cup
swings on a swing.}, Sentence 2: \textit{A child is making a ridiculous
face with an open mouth.}, Rating: 4, both sentences have 1 person.
Human annotators give a high similarity score of 4, even though the numbers are exactly the same.

Considering the condition \textbf{\textit{Age of person}}:

In the instance that Sentence 1: \textit{The boy on the bike is wearing safety glasses and a red helmet} and Sentence 2 is: \textit{A man dressed in bicycle gear is riding through a course}, the rating is 1. 
The perceived age difference between “boy” and “man” leads to a low similarity rating. 
Some annotators may weigh age references heavily when evaluating similarity.

In contrast, in the instance that Sentence 1: \textit{Two images show a man reaching out to hit a tennis ball with a racket} and Sentence 2 is: \textit{A boy in black shorts jumps and holds his tennis racket out in front of him}, the rating is 3. 
While the age difference between “man” and “boy” still exists, annotators give a moderate similarity score.

In another instance that Sentence 1: \textit{A very happy child sits on a chair
on top of some rocks.} and Sentence 2 is: \textit{A child is bouncing on a trampo-
line that is near a house}, the rating is 3.
Both sentences have description about the "child", which should be a higher similarity score of 4.
At least, the label should be different with the previous example which compares the age of "man" and "child".

Considering the condition \textbf{\textit{Gnender of person}}:

In the instance that Sentence 1: \textit{A man in a red and yellow outfit is riding a bicycle on one wheel} and Sentence 2: \textit{A woman is riding a bike with a basket of flowers}, the rating is 1.
Some annotators view gender as a central feature for this condition, leading to a low similarity rating despite shared activity.

However, in the instance that Sentence 1: \textit{A woman with a red scarf around her neck is smiling} and Sentence 2: \textit{A man in a black hat looks very happy}, the rating is 4.
Even though the genders differ, the facial expressions and emotional tone are similar, suggesting that some annotators focus more on affective similarity than gender cues, which is inaccurate.

\begin{table*}[ht!]
\centering
\begin{tabular}{lll}
\textbf{Condition issue} & \textbf{Before} & \textbf{After} \\
\toprule
Ambiguous Condition & The animal. & type of animal \\
 & The sport. & presence of vehicles\\ 
 \midrule
Unrelated Condition & The name of the game. & type of sport \\ 
\midrule
Inconsistent Phrasing Style & What the person is holding. & object being held \\
\midrule
Varying Granularity & The setting. & urban environment \\
& Specific areas of the home. & areas of home \\
\midrule
Verbose Expression & If a tv is present. & presence of television \\
\midrule
Grammatical Issue & The food with plate. & food on plate \\
& The the size of the room. & size of room \\ \bottomrule
\end{tabular}
\caption{Examples of conditions before and after using our condition modification prompt.}
\label{tab:condition-before-after}
\end{table*}

\begin{table*}[ht!]
\centering
\begin{tabular}{p{4.6cm} p{4.6cm} p{2.7cm} c c}
\textbf{Sentence 1} & \textbf{Sentence 2} & \textbf{Condition} & \textbf{Before} & \textbf{After} \\
\toprule
A room that has white walls and a window shade up has a double unmade bed on the floor. & A bed appears to have nothing else on it except two pillow in a bedroom. & type of room & 2 & 4\\
\hline
A deep dish pizza in a metal pan topped with several kinds of toppings. & The margarita pizza is on a plate, and ready to be cut and served. & type of pizza &  5 & 3 \\
\hline
Older men sitting on wooden benches on a sidewalk together, with scooters parked in the street and stores across the street. & There are people looking at a booth and a woman and man in a wheelchair on the sidewalk. & gender of people & 5 & 3 \\
\hline
a man sitting on a couch with a silver laptop in a living room & A computer desk topped with a monitor and a keyboard next to a  mouse. & number of people & 4 & 1 \\
\hline
A person flying a kite at the beach while two others walk past him & Three people standing on the shore of a sandy beach in front of waves & action of people & 5 & 3 \\
\hline
A colorful purple airplane sits on the runway with a darkened sky in the background. & A white and gray passenger plane has just landed or is about to take off. & type of vehicle & 2 & 4 \\
\hline
Two elephants are bathing in deep water as a person sits on one of their backs. & A group of people stand on the shore while watching an elephant in the water. & name of animal & 2 & 4 \\
\bottomrule
\end{tabular}
\caption{Examples of ratings with modified condition before and after using our re-annotation prompt.}
\label{tab:label-before-after}
\end{table*}

In the instance that Sentence 1: \textit{A little girl is brushing her teeth
in a bathroom.} and Sentence 2: \textit{A woman is brushing her teeth in
a bathroom mirror.}, the rating is 1.
The gender is both sentences is female.
Human annotators should not give a dissimilar score based on gender. 
When gender information matches across two sentences, it should not contribute to a higher dissimilarity rating.

In the instance that Sentence 1: \textit{A man is skateboarding on the
sidewalk.} and Sentence 2: \textit{A girl is rollerblading on a path. }, the rating is 4.
The gender is male in Sentence 1, but the gender is female in Sentence 2.
Humman annotators should not give a high similarity score of 4 to this mismatching gender information.

\section{Overlapping Statistics between original training and test sets}
\label{overlapping}
The overlaps between the original training and test sets by \citet{Deshpande:2023} are counted across the following five types:
\begin{itemize}
    \item Sentence only \\
    The same sentence appears, but possibly with different conditions.\\
    Overlap count: 1,196 \\
    Test side: $27.08\%$ of sentences overlap

    \item Condition only \\
    The same condition text appears, but possibly paired with different sentences.\\
    Overlap count: 804 \\
    Test side: $40.75\%$ of conditions overlap

    \item Single Sentence with Condition \\
    A single sentence–condition pair is repeated.\\
    Overlap count: 185 \\
    Test side: $1.96\%$ overlap

    \item Sentence pair (order-insensitive) \\
    The same pair of sentences appears (regardless of order).\\
    Overlap count: 9 \\
    Test side: $0.38\%$ overlap

    \item Sentence pair with Condition \\
    A full instance (two sentences with a condition) is duplicated.\\
    Overlap count: 2 \\
    Test side: $0.042\%$ overlap
\end{itemize}

Over one quarter of test sentences and over two-fifths of test conditions are also seen in the training set.
Such overlaps may lead to overestimated performance for language models.

\section{Prompt Used for Modifying the Conditions}
\label{cond-modification}

\autoref{fig:condition_prompt} shows the full prompt for condition modification.
\autoref{tab:condition-before-after} provides examples of how our prompt effectively refines various types of problematic conditions.

\section{Prompt Used for Similarity Annotations}
\label{label-reanno}

\autoref{fig:few_shot_prompt} shows the complete prompt for assigning similarity ratings using \acp{LLM}.
\autoref{tab:label-before-after} provides examples of the original and our re-annotated ratings, showing the improvement in the accuracy of \ac{C-STS} scores.
Selected examples are based on the conditions of the same semantic focus (conditions modified only with stopword removal).

\section{Evaluating the Averaging Method}
\label{voting}

\autoref{tab:voting} reports the average performance across different rating aggregation strategies.
We use Train-Mod training set with ratings as shown in the table.
We use \ac{NV} to first generate condition-aware sentence embeddings and then train the supervised multi-head non-linear projection as described in \autoref{sec:exp}.
Embeddings are evaluated on the ReTest-Mod test set.
The projection model is fixed with a hidden dimensionality of 1024, output dimensionality of 512, and a dropout rate of 0.1. 
Results show that combining human ratings with annotations from both \acp{LLM} yields the highest performance.

\begin{table}[ht]
\centering
\begin{tabular}{l r}
\textbf{Rating Data} & \textbf{Spearman} \\
\toprule
$y^{\text{GPT-4o}}$  & 70.88 \\
$y^{\text{Claude}}$ & 71.95 \\
V($y^{\text{GPT-4o}}$ + $y^{\text{Claude}}$) & 72.21 \\
V($y^{\text{GPT-4o}}$ + $y^{\text{human}}$) & 71.11 \\
V($y^{\text{Claude}}$ + $y^{\text{human}}$) & 72.74 \\
V($y^{\text{human}}$ + $y^{\text{GPT-4o}}$ + $y^{\text{Claude}}$) & \textbf{73.10} \\
\bottomrule
\end{tabular}
\caption{Average Spearman Correlation based on rating data across different aggregation strategies. V() denotes taking the arithmetic mean and rounding to the nearest integer.}
\label{tab:voting}
\end{table}

\section{Supervised Non-Linear Projection}
\label{multi-head}

The supervised non-linear projections are proposed by \citet{zhang2025caseconditionawaresentence}.
These supervised models are Siamese bi-encoders tailored for the \ac{C-STS} task which have proven high performance \cite{Deshpande:2023, Hyper-CL}.
Each model takes as input two condition-aware embeddings corresponding to sentence 1 and sentence 2 with the condition, respectively.

\citet{zhang2025caseconditionawaresentence} propose that input condition-aware sentence embeddings are generated from \ac{LLM}-based models, using the prompt ``Retrieve semantically similar texts to the [CONDITION], given the Sentence: [SENTENCE].''
They show that the \ac{LLM}-based embeddings work better than the \ac{MLM}-based embeddings.
To improve the condition-specific relevance, a post-processing step of subtracting the corresponding embeddings of the conditions is applied after generating the condition-aware sentence embeddings.
Here, the embeddings of the conditions are generated using the prompt ``Retrieve semantically similar texts to a given Sentence: [CONDITION].''

Denote the resulting \ac{LLM}-generated condition-aware sentence embeddings by $\mathbf{e}_1, \mathbf{e}_2$ for each instance. 
The \textbf{Supervised Non-Linear Projection} (SNPro) is defined as $f(\cdot)$, a two-layer feed-forward network with ReLU activations and dropout.  
The final projected embeddings are obtained as
\[
\mathbf{z}_i = f(\mathbf{e}_i), \quad i \in \{1,2\}.
\]

Hyperparameters are tuned on our validation set ReVal-Mod.
We fix the batch size to 512, the dropout rate to 0.15 and the learning rate to $10^{-3}$.
We select the output dimensionality of 512.

\begin{table}[ht]
    \centering
    \renewcommand{\arraystretch}{1.1} 
    \setlength{\tabcolsep}{4pt} 
    \resizebox{\linewidth}{!}{ 
    \begin{tabular}{lcc}
        \textbf{Model} &  \textbf{Non-linear \ac{FFN}} & \textbf{Linear \ac{FFN}}  \\
        \toprule
        \acs{NV} & 69.30 & \textbf{69.95}  \\
        \acs{SFR} & \textbf{62.85} & 59.22 \\
        \acs{GTE} & \textbf{64.16} & 56.10  \\
        \acs{E5} & \textbf{62.12} & 47.03 \\
        \acs{SimCSE-large} & \textbf{56.67} & 45.96\\
        \acs{SimCSE-base} & \textbf{56.60} & 39.54 \\
        \bottomrule
    \end{tabular}
    }
    \caption{Spearman correlation of embedding models based on supervised \acp{FFN} with reduced dimensionality $512$.}
    \label{tab:embedding_comparison}
\end{table}

\citet{zhang2025caseconditionawaresentence} found that \ac{LLM}-based models work better than \ac{MLM}-based models such as SimCSE for the \ac{C-STS} task.
Although a direct comparison with prior \ac{C-STS} methods is challenging due to issues in the test sets and lack of implementation details (e.g., \citet{L-CSTS} do not release their hyperparameters or test/validation splits), we include a comparison table to highlight the performance improvements achieved using the method proposed by \citet{zhang2025caseconditionawaresentence}.
\autoref{tab:embedding_comparison} shows the performance of different embedding models.
Three are LLM-based: 
\textit{NV-Embed-v2} (\textbf{\acs{NV}}), 
\textit{SFR-Embedding-Mistral} (\textbf{\acs{SFR}}), 
\textit{gte-Qwen2-7B-instruct} (\textbf{\acs{GTE}}).
Three are MLM-based: 
\textit{Multilingual-E5-large-instruct} (\textbf{\acs{E5}}), \textit{sup-simcse-roberta-large} (\textbf{\acs{SimCSE-large}}), and \textit{sup-simcse-bert-base-uncased} (\textbf{\acs{SimCSE-base}}).
\footnote{All models are available at \url{https://huggingface.co/spaces/mteb/leaderboard} and \url{https://huggingface.co/princeton-nlp}}
\ac{NV} achieves the highest Spearman correlation, significantly outperforming all other models.
Therefore, we select \ac{NV} as the base model for evaluating dataset cleansing effectiveness in our study.

\clearpage

\begin{figure*}[ht]
\centering
\begin{tcolorbox}[colback=gray!5!white,
                  colframe=gray!75!black,
                  width=\textwidth,
                  boxrule=0.5pt,
                  arc=4pt,
                  auto outer arc,
                  left=4pt,
                  right=4pt,
                  breakable,
                  enhanced jigsaw,
                  fontupper=\ttfamily,
                  before upper=\setlength{\parskip}{0pt}\setlength{\parindent}{0pt}]
\footnotesize
This is a Conditional STS task: Evaluate the similarity between the two sentences, with respect to the condition.

Sentence pair has a label (score) between 1 and 5 as follows:
Assign the pair a score between 1 and 5 as follows: \\
1. The two sentences are completely dissimilar with respect to the condition.\\
2. The two sentences are dissimilar, but are on a similar topic with respect to the condition.\\
3. The two sentences are roughly equivalent, but some important information differs or is missing with respect to the condition.\\
4. The two sentences are mostly equivalent, but some unimportant details differ with respect to the condition.\\
5. The two sentences are completely equivalent with respect to the condition.\\

Check and modify the provided condition if it is inaccurate or ambiguous, following these guidelines strictly:\\
* Conditions must be clear and specific. (e.g., instead of "animal", specify clearly such as "species of animal".)\\
* Remove stopword from conditions (e.g., "the").\\
* Conditions must accurately match human-annotated labels.\\
* Provide conditions concisely, without context-specific details. Good examples: color of clothing, type of event, intention of travel.\\
* Do NOT overly specify the condition more narrowly than the original meaning.\\

Return a JSON object with two fields:\\
\texttt{improved\_condition}: the improved condition,\\
\texttt{justification}: a single sentence explaining why you update the condition.\\
Give empty str this if only stopword 'the' is removed. 

\end{tcolorbox}
\caption{Prompt for modifying conditions}
\label{fig:condition_prompt}
\end{figure*}

\clearpage

\begin{figure*}[ht]
\centering
\begin{tcolorbox}[colback=gray!5!white,
                  colframe=gray!75!black,
                  width=\textwidth,
                  boxrule=0.5pt,
                  arc=4pt,
                  auto outer arc,
                  left=4pt,
                  right=4pt,
                  breakable,
                  enhanced jigsaw,
                  fontupper=\ttfamily,
                  before upper=\setlength{\parskip}{0pt}\setlength{\parindent}{0pt}]
\footnotesize
Definition: Evaluate the similarity between the two sentences, with respect to the condition. 
Assign the pair a score between 1 and 5 as follows: \\
1. The two sentences are completely dissimilar with respect to the condition.\\
2. The two sentences are dissimilar, but are on a similar topic with respect to the condition.\\
3. The two sentences are roughly equivalent, but some important information differs or is missing with respect to the condition.\\
4. The two sentences are mostly equivalent, but some unimportant details differ with respect to the condition.\\
5. The two sentences are completely equivalent with respect to the condition.\\

Evaluate the similarity for condition type "number of", following these guidelines strictly: \\
* Numbers need to be counted explicitly (e.g., “a man and a woman” → 2 people) \\
* If the two sentences mention the same number of entities → Label = 5 \\
* If the numbers differ → Label = 1 \\
* If no explicit number, follow the definition above and judge based on approximate quantity (e.g., "many" vs "a few"). \\

Return a JSON object with two fields:\\
"rating": the similarity rating (between 1 to 5 as defined above),\\
"justification": a single sentence explaining why you gave that similarity rating.\\

Do not return anything else other than this JSON object.  \\
Do not use code blocks. \\

\#\# Example 1\\
Sentence1: A close up of a giraffe laying on a ground near many large rocks.\\
Sentence2: A giraffe reaches up his head on a ledge high up on a rock.\\
Condition: animal's posture\\
\{"rating": 1, "justification": "In Sentence1 the giraffe is lying down, while in Sentence2 the giraffe is stretching its head upward."\}\\

\#\# Example 2\\
Sentence1: This bathroom stall has toilet tissue on the floor while the toilet is raised.\\
Sentence2: A full trashcan is beside the commode in a public restroom toilet that needs to be cleaned.\\
Condition: location of trash\\
\{"rating": 2, "justification": "Sentence2 does not clearly state that there is any trash outside the trashcan."\}\\

\#\# Example 3\\
Sentence1: A large red and blue boat sitting on top of a lake next to other boats.\\
Sentence2: Part of a ship sits in the shallow end of the bay next to a city.\\
Condition: body of water type\\
\{"rating": 3, "justification": "The two sentences mention lake and bay and are roughly equivalent, but Sentence2 does not clarify whether it is a bay within a lake."\}\\

\#\# Example 4\\
Sentence1: A monkey mug in front of a computer with a stuffed penguin beside it.\\
Sentence2: A laptop computer sitting on top of a table next to two computer monitors.\\
Condition: name of the device\\
\{"rating": 4, "justification": "Both sentences mention computers, but Sentence1 does not specify the type, while Sentence2 explicitly mentions a laptop."\}\\

\#\# Example 5\\
Sentence1: This bathroom stall has toilet tissue on the floor while the toilet is raised.\\
Sentence2: A full trashcan is beside the commode in a public restroom toilet that needs to be cleaned.\\
Condition: room function\\
\{"rating": 5, "justification": "Both sentences describe a room functioning as a restroom or toilet."\}
\end{tcolorbox}
\caption{Few-shot prompt for conditional sentence similarity annotation}
\label{fig:few_shot_prompt}
\end{figure*}

\end{document}